\documentclass[letterpaper, 10 pt, conference]{ieeeconf}  

\IEEEoverridecommandlockouts                              

\usepackage{amsmath} 
\usepackage{amssymb}  
\usepackage{graphicx}
\usepackage{caption}
\usepackage{booktabs}
\usepackage{float}
\usepackage{cite}
\usepackage{multirow}
\usepackage{flushend}

\title{\LARGE \bf
Reinforcement Learning for Wheeled Mobility on\\ Vertically Challenging Terrain
}

\author{Tong Xu, Chenhui Pan, and Xuesu Xiao
\thanks{All authors are with the Department of Computer Science, George Mason University {\tt\small \{txu25, cpan7, xiao\}@gmu.edu}}
}

\begin{document}

\maketitle
\thispagestyle{empty}
\pagestyle{empty}

\begin{abstract}

Off-road navigation on vertically challenging terrain, involving steep slopes and rugged boulders, presents significant challenges for wheeled robots both at the planning level to achieve smooth collision-free trajectories and at the control level to avoid rolling over or getting stuck. Considering the complex model of wheel-terrain interactions, we develop an end-to-end Reinforcement Learning (RL) system for an autonomous vehicle to learn wheeled mobility through simulated trial-and-error experiences. Using a custom-designed simulator built on the Chrono multi-physics engine, our approach leverages Proximal Policy Optimization (PPO) and a terrain difficulty curriculum to refine a policy based on a reward function to encourage progress towards the goal and penalize excessive roll and pitch angles, which circumvents the need of complex and expensive kinodynamic modeling, planning, and control. Additionally, we present experimental results in the simulator and deploy our approach on a physical Verti-4-Wheeler (V4W) platform, demonstrating that RL can equip conventional wheeled robots with previously unrealized potential of navigating vertically challenging terrain.

\end{abstract}
\section{INTRODUCTION}

\begin{figure}[h]
    \centering
    \includegraphics[width=\columnwidth]{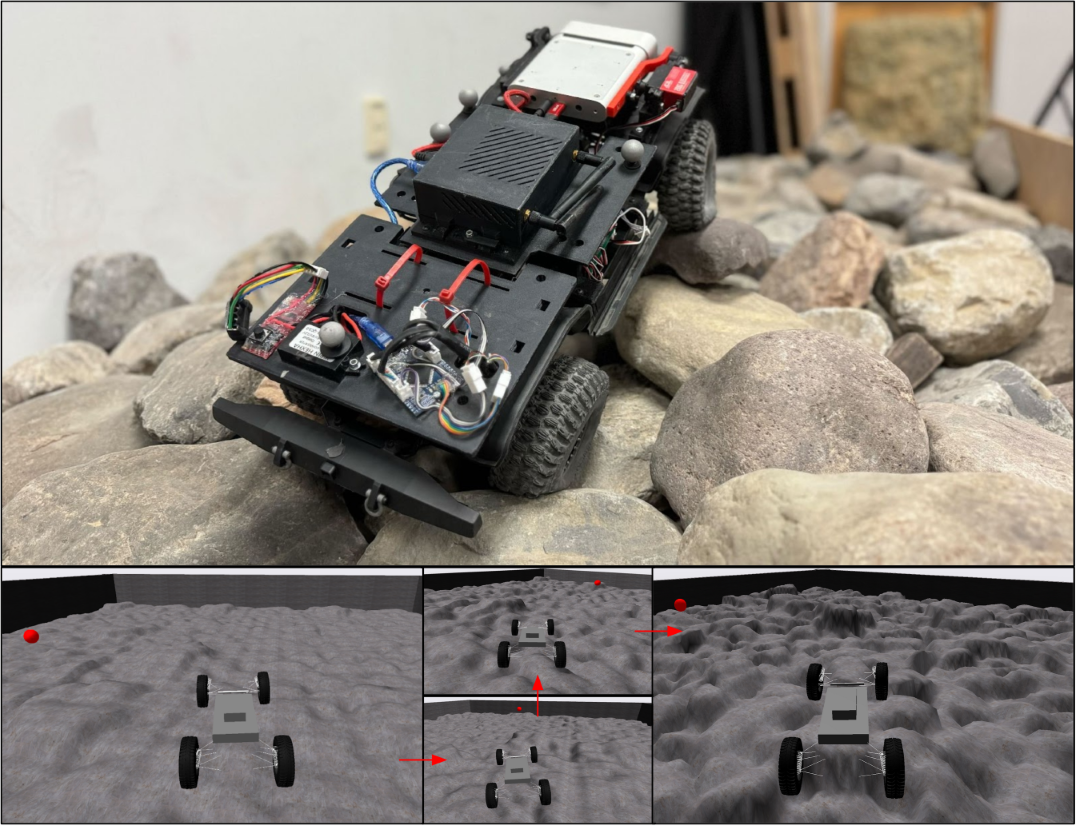}
    \caption{\texttt{VW-Chrono}: Simulator for Wheeled Mobility on Vertically Challenging Terrain with Increasing Difficulty (Lower Left to Lower Right).}
    \label{fig::VW-chrono}
\end{figure}

Autonomous off-road navigation has various safety, security, and rescue applications, such as search and rescue missions in hazardous or difficult-to-reach environments and scientific exploration in remote deserts or extraterrestrial planets~\cite{teji2023survey}. One particular thrust in this area of research is the development of widely available wheeled robots capable of navigating vertically challenging terrain (e.g., steep slopes, rocky outcroppings, and uneven surfaces, Fig.~\ref{fig::VW-chrono} top)~\cite{datar2023toward}. Achieving reliable and robust mobility in these environments is challenging due to the intricate nature of the terrain, the complex vehicle-terrain interactions, the adverse impact caused by gravity, and the potential deformation of the vehicle chassis.

Despite advancements in classical planning and control for off-road navigation, significant challenges remain. One major issue is the difficulty in precisely modeling vehicle-terrain interactions, which are highly variable and unpredictable in off-road, especially vertically challenge, environments. Implementing a high-precision kinodynamics or vehicle-terrain interaction model within a sampling-based motion planner can consume excessive computational resources onboard a mobile robot. Additionally, errors in these models can cascade into subsequent planning and control processes, leading to suboptimal performance. Furthermore, integrating multiple sensors and control algorithms increases system complexity and makes it challenging to generalize and scale across different terrain and applications.

To address these challenges, an increasing number of research efforts have introduced RL methods into off-road navigation. RL algorithms, such as Proximal Policy Optimization (PPO)~\cite{schulman2017proximal}, enables autonomous vehicles to learn and adapt to complex terrain through trial and error in simulation, without the need for costly real-world or expert demonstration data. Learning from a high-precision physics model in a simulator with RL in advance can also alleviate onboard computation during deployment. 

To advance off-road navigation solutions for wheeled robots on vertically challenging terrain using RL, we first develop a novel simulation environment developed within the Chrono multi-physics simulation engine~\cite{tasora2016chrono}. This simulator allows RL for wheeled robots to navigate vertically challenging terrain, with subsequent deployment onto a physical Verti-4-Wheeler (V4W)~\cite{datar2023toward}. We compare our navigation policy learned through PPO against an optimistic planner baseline and a classical planner with elevation approach, which shows the advantage of the RL-learned mobility. 
In summary, our contributions are outlined as follows:

\begin{figure*}[h]
    \centering
    \includegraphics[width=\textwidth]{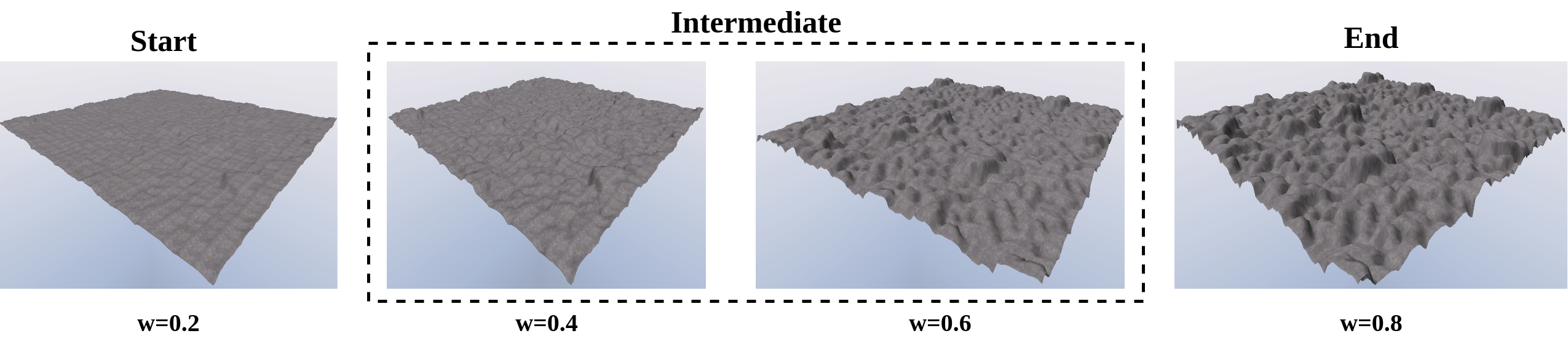}
    \caption{Increasing Mobility Difficulty of Vertically Challenging Terrain by Interpolating Start and End with Weight $w$.}
    \label{CL-BMP}
\end{figure*}

\begin{itemize}
    \item We create a simulator for wheeled mobility on vertically challenging terrain, \texttt{VW-Chrono} (Fig.~\ref{fig::VW-chrono} bottom), that procedurally generates four levels of increasing mobility difficulty to incorporate the principle of curriculum learning~\cite{narvekar2020curriculum}.
    \item We utilize PPO~\cite{schulman2017proximal} combined with the Sliced-Wasserstein Autoencoder (SWAE) structure~\cite{kolouri2018sliced} to efficiently learn wheeled mobility in \texttt{VW-Chrono}.
    \item We present a comparative study between our RL-learned mobility and two baselines for autonomously driving wheeled robots over vertically challenging terrain. 
\end{itemize}
\section{Related Work}
In this section, we provide a comprehensive review of related work on off-road mobility, focusing on both classical approaches and recent advances in data-driven methods. 

Off-road mobility presents significant challenges for autonomous robots due to the complexity and variability of unstructured terrains. Classical approaches have traditionally addressed these challenges by employing hand-crafted methodologies for perception~\cite{lu2014layered}, planning~\cite{rastgoftar2018data}, modeling~\cite{he2019review}, and control~\cite{williams2016aggressive}. These techniques often rely on heuristics and extensive domain expertise to handle environmental variations. While effective in controlled scenarios, these classical methods suffer from several notable limitations: they require significant engineering effort, are susceptible to cascading errors from upstream perception and planning modules, and struggle to adapt effectively to novel or unforeseen environments~\cite{thrun2006stanley}.

To overcome the shortcomings of traditional methods, data-driven approaches for off-road mobility have emerged as promising alternatives~\cite{xiao2022motion}. These methods leverage advances in machine learning to directly learn complex behaviors from data, offering adaptability in environments that are too intricate for manual engineering~\cite{xiao2022motion}. Among these methods, end-to-end learning of control policies has been explored extensively, where imitation learning~\cite{pan2020imitation} and reinforcement learning (RL) are used to learn robust navigation strategies from either expert demonstrations or trial-and-error interactions. Moreover, learning-based semantic perception methods have been employed to provide high-level scene understanding and terrain classification for improved mobility~\cite{manduchi2005obstacle, maturana2018real, shaban2022semantic, meng2023terrainnet, viswanath2021offseg, sikand2022visual}.

In addition to perception, recent efforts have also focused on learning kinodynamic models~\cite{xiao2021learning, karnan2022vi, atreya2022high, datar2024terrain, datar2023learning, pokhrel2024cahsor, maheshwari2023piaug, nazeri2024vertiencoder} that better capture the physical interactions between the robot and varying terrain types. Parameter adaptation approaches~\cite{xiao2020appld, wang2021appli, wang2021apple, xu2021applr, xiao2022appl} have also been proposed to adjust system parameters on-the-fly based on perception feedback, providing greater robustness to environmental changes. Furthermore, learning-based cost function optimization~\cite{sivaprakasam2021improving, dashora2022hybrid, cai2022risk, castro2023traversability, cai2024evora, cai2024pietra, seo2023learning, jung2024v, xiao2022learning, pan2024traverse} has contributed to improved decision-making by enabling more nuanced and context-aware trajectory planning.

Despite their promise, data-driven approaches face notable challenges. Specifically, RL~\cite{xu2023benchmarking, xu2021machine, xu2024dexterous} and imitation learning~\cite{karnan2022socially, nguyen2023toward, karnan2022voila} methods tend to be data-intensive, often requiring either millions of trial-and-error iterations or substantial expert-provided labeled datasets~\cite{xiao2021toward, xiao2021agile, wang2021agile, ghani2024dyna} for effective policy learning. Furthermore, ensuring generalization of learned models to diverse, unseen environments remains a critical open question. One potential solution lies in curriculum learning, where a sequence of progressively challenging tasks is presented to the agent~\cite{narvekar2020curriculum, wang2024grounded}. This strategy has shown potential for improving both sample efficiency and robustness of learned policies, thereby facilitating better generalization across different deployment settings.

\section{METHOD}

In this section, we present the design of  \texttt{VW-Chrono} and its OpenAI Gym environment. We introduce our RL problem and training for wheeled mobility on vertically challenging terrain, as well as our SWAE-based elevation map encoder.

\subsection{\texttt{VW-Chrono}}

To ensure the simulated vertically challenging terrain resemble the real world, we first utilize our physical V4W to collect elevation map data on a custom-built indoor testbed designed for vertically challenging terrain. This testbed includes hundreds of rocks and boulders, averaging 30cm in size (matching the scale of the V4W), which are randomly laid out and stacked on a 3.1$\times$1.3m test course. The highest elevation of the test course can reach up to 0.5m, more than twice the height of the vehicle (Fig.~\ref{fig::VW-chrono} top). We create a grayscale Bitmap image (BMP) with the collected data to represent terrain elevation~\cite{miki2022elevation}. In the Chrono multi-physics simulation engine, a triangular mesh is generated by assigning a vertex to each pixel of the BMP image. The mesh is then horizontally to match the given extents and expanded vertically to align with the specified range. This ensures that the darkest pixel aligns with the minimum height and the lightest pixel corresponds to the maximum height (Fig.~\ref{fig::VW-chrono} bottom). 

To create vertically challenging environments with different difficulty levels as shown in Fig.~\ref{CL-BMP}, we create a sequence of elevation maps by linearly interpolating between a starting map $I_0$ (flat terrain) and an ending map $I_N$ (rugged terrain) using a weighted average. The intermediate image $I_k$ at stage $k$ out of $N$ stages can be calculated using the following equation:
\begin{equation}
I_k = (1 - \frac{k}{N}) I_0 + \frac{k}{N} I_N,\quad\forall k\in\{0, 1, ..., N\}.
\label{eqn::CL}
\end{equation}
In Eqn.~\eqref{eqn::CL}, the term \( \displaystyle \frac{k}{N} \) is used to define the interpolation weight $w$ in Fig.~\ref{CL-BMP}. This approach is based on the principle of curriculum learning, which posits that models can learn more effectively and efficiently when tasks are introduced in a structured, incremental manner, starting with simpler tasks and gradually moving to more complex ones.

\subsection{RL Problem Formulation}

We employ RL to train a policy that receives environmental inputs and generates actions to drive the robot through vertically challenging terrain, avoiding getting stuck and rolling over while moving toward a designated goal. 

A regular Markov Decision Process (MDP) can be defined by a tuple $(\mathcal{S}, \mathcal{A}, \mathcal{T}, \gamma, \mathcal{R})$, including state, action, state transition, discount factor and reward. The goal is to learn a policy $\pi: \mathcal{S} \to \mathcal{A}$ to maximize the expected cumulative reward over the task horizon $T$, i.e.,
\begin{equation}
\max_\pi \ \mathbb{E}_{a_t \sim \pi(\cdot | s_t)} \left[ \sum_{t=0}^T \gamma^t R_t \right], 
\label{eqn::reward}
\end{equation}
where $a_t \in \mathcal{A}$ and $s_t \in \mathcal{S}$ are the action and state of the system at each step. To learn wheeled mobility for vertically challenging terrain, the following design choices are made: 

\subsubsection{State Space} The inputs to our RL policy include angular difference between the vehicle and goal heading (in radian), current vehicle velocity (in m/s), and cropped elevation map centered at and aligned with the vehicle. We use a Sliced-Wasserstein Autoencoder (SWAE) to reduce elevation map dimensionality and utilize the latent vector to preserve original elevation information. After SWAE pretraining, we freeze the parameters of the encoder during RL training. 

\subsubsection{Action Space} The RL policy's outputs include desired linear speed and steering angle, instead of raw throttle and steering commands, in order to improve learning efficiency. A PID controller controls the throttle and steering commands to achieve the desired linear speed and steering angle. 

\subsubsection{Policy Model Architecture} We choose PPO~\cite{schulman2017proximal} as the RL algorithm considering our continuous action space. PPO iteratively collects data through interactions with the environment and updates the policy to maximize the expected cumulative reward (Eqn.~\eqref{eqn::reward}). Unlike traditional methods, PPO employs a clipped surrogate objective to constrain policy updates, preventing significant deviations that could lead to instability. By balancing the exploration-exploitation trade-off with a proximal threshold, PPO continually improves the policy while ensuring stability.

\begin{table*}[t]
\centering
\caption{Experiment Results of RL and Baselines}
\renewcommand{\arraystretch}{1.3}
\begin{tabular}{ccccc}
\toprule[1pt]
Approach & Stage 1 & Stage 2 & Stage 3 & Stage 4 \\
\midrule
RL & 
\begin{tabular}[c]{@{}c@{}}
\textbf{25/25}, 5.75s,\\
\textbf{1.19$^{\circ}$/1.26$^{\circ}$} 
\end{tabular}
& 
\begin{tabular}[c]{@{}c@{}}
\textbf{25/25}, 5.23s,\\
\textbf{2.59$^{\circ}$}/2.30$^{\circ}$
\end{tabular}
& 
\begin{tabular}[c]{@{}c@{}}
\textbf{20/25}, \textbf{5.35s},\\
\textbf{3.55$^{\circ}$/2.23$^{\circ}$} 
\end{tabular}
& 
\begin{tabular}[c]{@{}c@{}}
\textbf{15/25}, 6.21s,\\
\textbf{5.58$^{\circ}$/3.82$^{\circ}$}
\end{tabular} 
\\
Optimistic Planner & 
\begin{tabular}[c]{@{}c@{}}
\textbf{25/25}, \textbf{4.65s},\\
1.36$^{\circ}$/1.50$^{\circ}$ 
\end{tabular}
& 
\begin{tabular}[c]{@{}c@{}}
\textbf{25/25}, \textbf{4.82s},\\
2.63$^{\circ}$/2.00$^{\circ}$ 
\end{tabular}
& 
\begin{tabular}[c]{@{}c@{}}
17/25, 5.46s,\\
4.32$^{\circ}$/2.97$^{\circ}$ 
\end{tabular}
& 
\begin{tabular}[c]{@{}c@{}}
10/25, \textbf{5.68s},\\
7.11$^{\circ}$/4.11$^{\circ}$ 
\end{tabular} 
\\
Naive Planner & 
\begin{tabular}[c]{@{}c@{}}
\textbf{25/25}, 5.39s,\\
1.31$^{\circ}$/1.37$^{\circ}$
\end{tabular}
& 
\begin{tabular}[c]{@{}c@{}}
\textbf{25/25}, 5.18s,\\
2.80$^{\circ}$/\textbf{1.99}$^{\circ}$ 
\end{tabular}
& 
\begin{tabular}[c]{@{}c@{}}
\textbf{20/25}, 6.20s,\\
4.90$^{\circ}$/2.76$^{\circ}$ 
\end{tabular}
& 
\begin{tabular}[c]{@{}c@{}}
12/25, 6.73s,\\
5.67$^{\circ}$/4.07$^{\circ}$ 
\end{tabular}
\\
\midrule
Best Reward Mean (RL) & 2860.9 & 2415.4 & 1393.3 & 739.5 
\\ 
\bottomrule[1pt]
\label{tab::results}
\end{tabular}
\end{table*}

\subsection{Sliced-Wasserstein Autoencoder (SWAE)}

We use SWAE as a feature extractor to reduce the dimension of the elevation map around the robot while preserving the original elevation information.
SWAE is a scalable generative model that captures the rich and often nonlinear distribution of high-dimensional data (e.g., images, videos, and audio). Learning such generative models involves minimizing a dissimilarity measure between the data distribution and the output distribution of the generative model, which essentially constitutes an optimal transport problem.





\subsection{Reward Design}

Our RL agent is trained using a reward function composed of three key terms. These terms are designed to incentivize the agent's movement toward the goal and prevent immobilization. The components of the reward function are:
\begin{equation}
R_t := R_{\text{progress}} + R_{\text{rollover}} + R_{\text{timeout}}
\end{equation}

\subsubsection{Progress Reward} This term promotes the agent's advancement toward the goal by providing positive rewards for progress made. Additionally, if the agent has not moved at least 1cm within 0.1 seconds, a penalty is applied: 
$$
R_\text{progress} = w_1 \cdot \Delta d - w_2 \cdot \mathbb{I}(\Delta d < 0.01),
$$
where $\Delta d$ is the distance moved towards the goal between the previous timestamp and the current timestamp, $\mathbb{I}()$ is an indicator function, and all different $w_i$ are weight terms.

\subsubsection{Rollover Penalty} To prevent the agent from rolling over, we penalize excessive roll and pitch angles:
\begin{equation}
R_{\text{rollover}} = -w_3 \cdot \sum_{i \in \{\text{roll}, \text{pitch}\}} \max(0, |\theta_i| - \alpha),
\end{equation}
where $\theta_{\text{roll}}$ and $\theta_{\text{pitch}}$ are the roll and pitch angles, respectively, $w_3$ is a weight term and $\alpha$ is a constant threshold angle.

\subsubsection{Timeout Penalty} For each episode, if a time limit  $T$ is reached before the robot reaching the goal, a fixed penalty $c$ is applied for the timeout, along with an additional penalty based on the remaining distance to the goal:
$$
R_\text{timeout} = - (w_4 \cdot d_\text{remaining} + c) \cdot \mathbb{I}(t \geq T),
$$
where $d_{\text{remaining}}$ is the remaining distance to the goal. 

Table \ref{tab::parameters} shows all hyper-parameters of our reward function.

\begin{table}[h]
\centering
\caption{Reward Weights}
\renewcommand{\arraystretch}{1.5}
\begin{tabular}{ccccccc}
\toprule[1pt]
$w_1$ & $w_2$ & $w_3$ & $w_4$ & $\alpha$ & $c$ & $T$ \\
\hline
50 & 10 & 20 & 10 & 30 & 100 & 15 \\
\bottomrule[1pt]
\label{tab::parameters}
\vspace{-20pt}
\end{tabular}
\end{table}
\section{RESULTS}

In this section, we present the experimental results of our RL system. compared against two baselines designed for vertically challenging terrain.

\subsection{Baselines} 
We design two baselines for our \texttt{VW-Chrono} simulation environment. 

\subsubsection{Optimistic Planner with flat-terrain assumption} The primary input for this controller is the angular difference between the vehicle's current heading and the desired heading towards the goal. By minimizing this angle difference, the planner guides the vehicle towards its target. 

\subsubsection{Naive Planner with elevation heuristic} The Optimistic Planner with flat-terrain assumption often struggles with steep slopes and rugged boulders, leading to the vehicle getting stuck. To enhance the planner's performance on challenging rock terrain, we employ a $64 \times 64$ cropped elevation map centered on the vehicle. From the front part of the vehicle, we evenly split the map into five regions and choose the most traversable direction: At each time step, we calculate the mean and variance of the elevation values of these five regions and select the region with the most similar mean and lowest variance as the driving direction, compared to the region of the same size centered at the vehicle. 

We utilize three metrics to compare results in Table \ref{tab::results}: 

\begin{enumerate}
    \item Number of successful trials (out of 25).
    \item Mean traversal time (of successful trails in seconds).
    \item Average roll/pitch angles (in degrees).
\end{enumerate}


\begin{figure*}[h]
    \centering
    \includegraphics[scale=0.4]{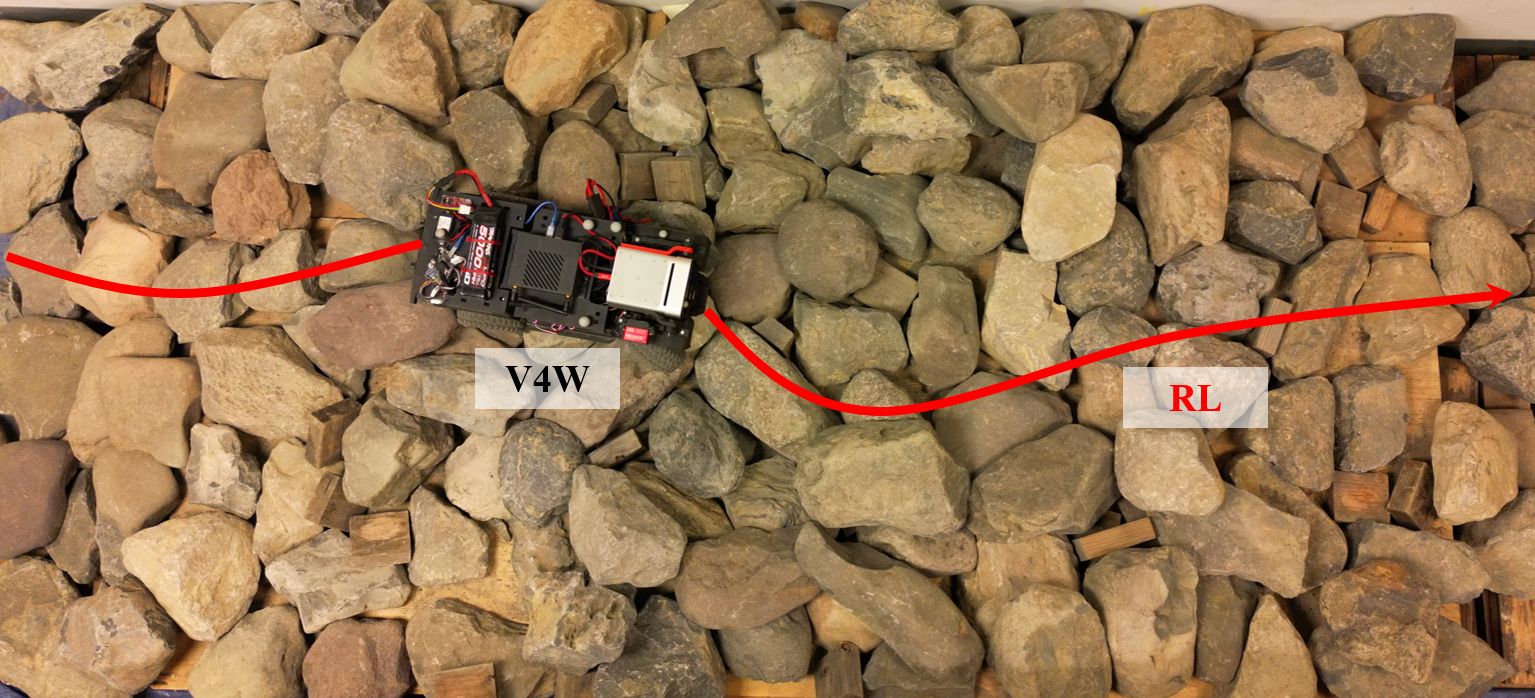}
    \caption{Custom-Built Testbed with V4W and an Example Trajectory by the RL Algorithm.}
    \label{RockTest}
\end{figure*}

\subsection{Simulation Results}

In \texttt{VW-Chrono}, we randomly set vehicle start and goal position on the testbed every time and test baselines against our RL system. We present our experiment results in Table \ref{tab::results}, where best results are shown in bold.
The four stages correspond to four increasing difficulty levels, 25 trials each. The RL method consistently achieves a high number of successful trials, particularly excelling in the earlier stages with a perfect success rate in Stages 1 and 2 and maintaining reasonable success rates in Stages 3, 4. However, in Stage 4, while the RL method achieves a success rate of 15 out of 25, it maintains the best roll/pitch stability compared to the Optimistic Planner and Naive Planner, indicating its effectiveness in handling complex terrain with slower and more cautious navigation.

The Optimistic Planner, while achieving the fastest traversal times, shows a decline in performance as the terrain difficulty increases, with a significant drop in the number of successful trials and increasing roll/pitch angles in Stages 3 and 4. This indicates that the Optimistic Planner, although efficient on less challenging terrain, struggles with stability and success in more complex environments.

The Naive Planner strikes a balance between speed and stability, with a high success rate and relatively low roll/pitch angles across all stages. It demonstrates superior performance over the Optimistic Planner in maintaining lower roll/pitch angles, particularly in the most difficult Stages 4. However, it still does not surpass the RL approach in terms of overall stability in those complex stages.

\subsection{Physical Demonstration}
We also deploy the RL policy learned in simulation on a physical V4W platform on a real-world rock testbed (Fig.~\ref{fig::VW-chrono} top). The robot is a four-wheeled platform based on an off-the-self, two-axle, four-wheel-drive, off-road vehicle from Traxxas. The onboard computation platform is a NVIDIA Jetson Xavier NX module. First, we place the V4W on flat terrain and specify a direction for it to follow. The RL policy successfully guides the V4W in the intended direction. Next, we introduce a large obstacle to assess the RL policy's performance. Finally, we test the V4W on the rock testbed and observe that the RL policy effectively enables the V4W to move toward its goal across the rocky terrain as shown in Fig.~\ref{RockTest}.
\section{CONCLUSION}

This paper presents a comprehensive RL system to unlock the previously unrealized potential of wheeled mobility on vertically challenging terrain. The \texttt{VW-Chrono} simulator can generate challenging terrain for future off-road navigation research with adjustable mobility difficulty levels. We utilize PPO as our RL algorithm based on a carefully designed reward structure. The experimental results confirm our hypothesis that conventional wheeled robots possess the mechanical capability to navigate vertically challenging terrain, which are normally considered as non-traversable obstacles, especially with the help of data-driven approaches. Furthermore, we demonstrate the feasibility of transferring RL-learned mobility from simulation to a physical robot, enabling it to navigate real-world vertically challenging terrain.

This paper opens up a new research direction aimed at achieving extreme off-road robot mobility using RL methods. One promising future research direction is to employ a teacher-student structure to automatically create different levels of terrain in an automatic curriculum learning setting to improve learning efficiency. 



\bibliographystyle{IEEEtran}
\bibliography{references}

\begin{thebibliography}{10}
\providecommand{\url}[1]{#1}
\csname url@samestyle\endcsname
\providecommand{\newblock}{\relax}
\providecommand{\bibinfo}[2]{#2}
\providecommand{\BIBentrySTDinterwordspacing}{\spaceskip=0pt\relax}
\providecommand{\BIBentryALTinterwordstretchfactor}{4}
\providecommand{\BIBentryALTinterwordspacing}{\spaceskip=\fontdimen2\font plus
\BIBentryALTinterwordstretchfactor\fontdimen3\font minus \fontdimen4\font\relax}
\providecommand{\BIBforeignlanguage}[2]{{%
\expandafter\ifx\csname l@#1\endcsname\relax
\typeout{** WARNING: IEEEtran.bst: No hyphenation pattern has been}%
\typeout{** loaded for the language `#1'. Using the pattern for}%
\typeout{** the default language instead.}%
\else
\language=\csname l@#1\endcsname
\fi
#2}}
\providecommand{\BIBdecl}{\relax}
\BIBdecl

\bibitem{teji2023survey}
M.~D. Teji, T.~Zou, and D.~S. Zeleke, ``A survey of off-road mobile robots: Slippage estimation, robot control, and sensing technology,'' \emph{Journal of Intelligent \& Robotic Systems}, vol. 109, no.~2, p.~38, 2023.

\bibitem{datar2023toward}
A.~Datar, C.~Pan, M.~Nazeri, and X.~Xiao, ``Toward wheeled mobility on vertically challenging terrain: Platforms, datasets, and algorithms,'' in \emph{2024 IEEE International Conference on Robotics and Automation (ICRA)}.\hskip 1em plus 0.5em minus 0.4em\relax IEEE, 2024.

\bibitem{schulman2017proximal}
J.~Schulman, F.~Wolski, P.~Dhariwal, A.~Radford, and O.~Klimov, ``Proximal policy optimization algorithms,'' \emph{arXiv preprint arXiv:1707.06347}, 2017.

\bibitem{tasora2016chrono}
A.~Tasora, R.~Serban, H.~Mazhar, A.~Pazouki, D.~Melanz, J.~Fleischmann, M.~Taylor, H.~Sugiyama, and D.~Negrut, ``Chrono: An open source multi-physics dynamics engine,'' in \emph{High Performance Computing in Science and Engineering: Second International Conference, HPCSE 2015, Sol{\'a}{\v{n}}, Czech Republic, May 25-28, 2015, Revised Selected Papers 2}.\hskip 1em plus 0.5em minus 0.4em\relax Springer, 2016, pp. 19--49.

\bibitem{narvekar2020curriculum}
S.~Narvekar, B.~Peng, M.~Leonetti, J.~Sinapov, M.~E. Taylor, and P.~Stone, ``Curriculum learning for reinforcement learning domains: A framework and survey,'' \emph{Journal of Machine Learning Research}, vol.~21, no. 181, pp. 1--50, 2020.

\bibitem{kolouri2018sliced}
S.~Kolouri, P.~E. Pope, C.~E. Martin, and G.~K. Rohde, ``Sliced-wasserstein autoencoder: An embarrassingly simple generative model,'' \emph{arXiv preprint arXiv:1804.01947}, 2018.

\bibitem{lu2014layered}
D.~V. Lu, D.~Hershberger, and W.~D. Smart, ``Layered costmaps for context-sensitive navigation,'' in \emph{2014 IEEE/RSJ International Conference on Intelligent Robots and Systems}.\hskip 1em plus 0.5em minus 0.4em\relax IEEE, 2014, pp. 709--715.

\bibitem{rastgoftar2018data}
H.~Rastgoftar, B.~Zhang, and E.~M. Atkins, ``A data-driven approach for autonomous motion planning and control in off-road driving scenarios,'' in \emph{2018 Annual american control conference (ACC)}.\hskip 1em plus 0.5em minus 0.4em\relax IEEE, 2018, pp. 5876--5883.

\bibitem{he2019review}
R.~He, C.~Sandu, A.~K. Khan, A.~G. Guthrie, P.~S. Els, and H.~A. Hamersma, ``Review of terramechanics models and their applicability to real-time applications,'' \emph{Journal of Terramechanics}, vol.~81, pp. 3--22, 2019.

\bibitem{williams2016aggressive}
G.~Williams, P.~Drews, B.~Goldfain, J.~M. Rehg, and E.~A. Theodorou, ``Aggressive driving with model predictive path integral control,'' in \emph{2016 IEEE International Conference on Robotics and Automation (ICRA)}.\hskip 1em plus 0.5em minus 0.4em\relax IEEE, 2016.

\bibitem{thrun2006stanley}
S.~Thrun, M.~Montemerlo, H.~Dahlkamp, D.~Stavens, A.~Aron, J.~Diebel, P.~Fong, J.~Gale, M.~Halpenny, G.~Hoffmann \emph{et~al.}, ``Stanley: The robot that won the darpa grand challenge,'' \emph{Journal of field Robotics}, vol.~23, no.~9, pp. 661--692, 2006.

\bibitem{xiao2022motion}
X.~Xiao, B.~Liu, G.~Warnell, and P.~Stone, ``Motion planning and control for mobile robot navigation using machine learning: a survey,'' \emph{Autonomous Robots}, vol.~46, no.~5, pp. 569--597, 2022.

\bibitem{pan2020imitation}
Y.~Pan, C.-A. Cheng, K.~Saigol, K.~Lee, X.~Yan, E.~A. Theodorou, and B.~Boots, ``Imitation learning for agile autonomous driving,'' \emph{The International Journal of Robotics Research}, vol.~39, no. 2-3, pp. 286--302, 2020.

\bibitem{manduchi2005obstacle}
R.~Manduchi, A.~Castano, A.~Talukder, and L.~Matthies, ``Obstacle detection and terrain classification for autonomous off-road navigation,'' \emph{Autonomous robots}, vol.~18, pp. 81--102, 2005.

\bibitem{maturana2018real}
D.~Maturana, P.-W. Chou, M.~Uenoyama, and S.~Scherer, ``Real-time semantic mapping for autonomous off-road navigation,'' in \emph{Field and Service Robotics}.\hskip 1em plus 0.5em minus 0.4em\relax Springer, 2018, pp. 335--350.

\bibitem{shaban2022semantic}
A.~Shaban, X.~Meng, J.~Lee, B.~Boots, and D.~Fox, ``Semantic terrain classification for off-road autonomous driving,'' in \emph{Conference on Robot Learning}.\hskip 1em plus 0.5em minus 0.4em\relax PMLR, 2022, pp. 619--629.

\bibitem{meng2023terrainnet}
X.~Meng, N.~Hatch, A.~Lambert, A.~Li, N.~Wagener, M.~Schmittle, J.~Lee, W.~Yuan, Z.~Chen, S.~Deng \emph{et~al.}, ``Terrainnet: Visual modeling of complex terrain for high-speed, off-road navigation,'' \emph{arXiv preprint arXiv:2303.15771}, 2023.

\bibitem{viswanath2021offseg}
K.~Viswanath, K.~Singh, P.~Jiang, P.~Sujit, and S.~Saripalli, ``Offseg: A semantic segmentation framework for off-road driving,'' in \emph{2021 IEEE 17th International Conference on Automation Science and Engineering (CASE)}.\hskip 1em plus 0.5em minus 0.4em\relax IEEE, 2021, pp. 354--359.

\bibitem{sikand2022visual}
K.~S. Sikand, S.~Rabiee, A.~Uccello, X.~Xiao, G.~Warnell, and J.~Biswas, ``Visual representation learning for preference-aware path planning,'' in \emph{2022 International Conference on Robotics and Automation (ICRA)}.\hskip 1em plus 0.5em minus 0.4em\relax IEEE, 2022, pp. 11\,303--11\,309.

\bibitem{xiao2021learning}
X.~Xiao, J.~Biswas, and P.~Stone, ``Learning inverse kinodynamics for accurate high-speed off-road navigation on unstructured terrain,'' \emph{IEEE Robotics and Automation Letters}, vol.~6, no.~3, pp. 6054--6060, 2021.

\bibitem{karnan2022vi}
H.~Karnan, K.~S. Sikand, P.~Atreya, S.~Rabiee, X.~Xiao, G.~Warnell, P.~Stone, and J.~Biswas, ``Vi-ikd: High-speed accurate off-road navigation using learned visual-inertial inverse kinodynamics,'' in \emph{2022 IEEE/RSJ International Conference on Intelligent Robots and Systems (IROS)}.\hskip 1em plus 0.5em minus 0.4em\relax IEEE, 2022, pp. 3294--3301.

\bibitem{atreya2022high}
P.~Atreya, H.~Karnan, K.~S. Sikand, X.~Xiao, S.~Rabiee, and J.~Biswas, ``High-speed accurate robot control using learned forward kinodynamics and non-linear least squares optimization,'' in \emph{2022 IEEE/RSJ International Conference on Intelligent Robots and Systems (IROS)}.\hskip 1em plus 0.5em minus 0.4em\relax IEEE, 2022, pp. 11\,789--11\,795.

\bibitem{datar2024terrain}
A.~Datar, C.~Pan, M.~Nazeri, A.~Pokhrel, and X.~Xiao, ``Terrain-attentive learning for efficient 6-dof kinodynamic modeling on vertically challenging terrain,'' in \emph{2024 IEEE/RSJ International Conference on Intelligent Robots and Systems (IROS)}.\hskip 1em plus 0.5em minus 0.4em\relax IEEE, 2024.

\bibitem{datar2023learning}
A.~Datar, C.~Pan, and X.~Xiao, ``Learning to model and plan for wheeled mobility on vertically challenging terrain,'' \emph{arXiv preprint arXiv:2306.11611}, 2023.

\bibitem{pokhrel2024cahsor}
A.~Pokhrel, A.~Datar, M.~Nazeri, and X.~Xiao, ``{CAHSOR}: Competence-aware high-speed off-road ground navigation in {SE} (3),'' \emph{IEEE Robotics and Automation Letters}, 2024.

\bibitem{maheshwari2023piaug}
P.~Maheshwari, W.~Wang, S.~Triest, M.~Sivaprakasam, S.~Aich, J.~G. Rogers~III, J.~M. Gregory, and S.~Scherer, ``Piaug--physics informed augmentation for learning vehicle dynamics for off-road navigation,'' \emph{arXiv preprint arXiv:2311.00815}, 2023.

\bibitem{nazeri2024vertiencoder}
M.~Nazeri, A.~Datar, A.~Pokhrel, C.~Pan, G.~Warnell, and X.~Xiao, ``Vertiencoder: Self-supervised kinodynamic representation learning on vertically challenging terrain,'' \emph{arXiv preprint arXiv:2409.11570}, 2024.

\bibitem{xiao2020appld}
X.~Xiao, B.~Liu, G.~Warnell, J.~Fink, and P.~Stone, ``Appld: Adaptive planner parameter learning from demonstration,'' \emph{IEEE Robotics and Automation Letters}, vol.~5, no.~3, pp. 4541--4547, 2020.

\bibitem{wang2021appli}
Z.~Wang, X.~Xiao, B.~Liu, G.~Warnell, and P.~Stone, ``Appli: Adaptive planner parameter learning from interventions,'' in \emph{2021 IEEE international conference on robotics and automation (ICRA)}.\hskip 1em plus 0.5em minus 0.4em\relax IEEE, 2021, pp. 6079--6085.

\bibitem{wang2021apple}
Z.~Wang, X.~Xiao, G.~Warnell, and P.~Stone, ``Apple: Adaptive planner parameter learning from evaluative feedback,'' \emph{IEEE Robotics and Automation Letters}, vol.~6, no.~4, pp. 7744--7749, 2021.

\bibitem{xu2021applr}
Z.~Xu, G.~Dhamankar, A.~Nair, X.~Xiao, G.~Warnell, B.~Liu, Z.~Wang, and P.~Stone, ``Applr: Adaptive planner parameter learning from reinforcement,'' in \emph{2021 IEEE international conference on robotics and automation (ICRA)}.\hskip 1em plus 0.5em minus 0.4em\relax IEEE, 2021, pp. 6086--6092.

\bibitem{xiao2022appl}
X.~Xiao, Z.~Wang, Z.~Xu, B.~Liu, G.~Warnell, G.~Dhamankar, A.~Nair, and P.~Stone, ``Appl: Adaptive planner parameter learning,'' \emph{Robotics and Autonomous Systems}, vol. 154, p. 104132, 2022.

\bibitem{sivaprakasam2021improving}
M.~Sivaprakasam, S.~Triest, W.~Wang, P.~Yin, and S.~Scherer, ``Improving off-road planning techniques with learned costs from physical interactions,'' in \emph{2021 IEEE International Conference on Robotics and Automation (ICRA)}.\hskip 1em plus 0.5em minus 0.4em\relax IEEE, 2021, pp. 4844--4850.

\bibitem{dashora2022hybrid}
N.~Dashora, D.~Shin, D.~Shah, H.~Leopold, D.~Fan, A.~Agha-Mohammadi, N.~Rhinehart, and S.~Levine, ``Hybrid imitative planning with geometric and predictive costs in off-road environments,'' in \emph{2022 International Conference on Robotics and Automation (ICRA)}.\hskip 1em plus 0.5em minus 0.4em\relax IEEE, 2022, pp. 4452--4458.

\bibitem{cai2022risk}
X.~Cai, M.~Everett, J.~Fink, and J.~P. How, ``Risk-aware off-road navigation via a learned speed distribution map,'' in \emph{2022 IEEE/RSJ International Conference on Intelligent Robots and Systems (IROS)}.\hskip 1em plus 0.5em minus 0.4em\relax IEEE, 2022, pp. 2931--2937.

\bibitem{castro2023traversability}
M.~G. Castro, S.~Triest, W.~Wang, J.~M. Gregory, F.~Sanchez, J.~G. Rogers, and S.~Scherer, ``How does it feel? self-supervised costmap learning for off-road vehicle traversability,'' in \emph{2023 IEEE International Conference on Robotics and Automation (ICRA)}, 2023, pp. 931--938.

\bibitem{cai2024evora}
X.~Cai, S.~Ancha, L.~Sharma, P.~R. Osteen, B.~Bucher, S.~Phillips, J.~Wang, M.~Everett, N.~Roy, and J.~P. How, ``{EVORA}: Deep evidential traversability learning for risk-aware off-road autonomy,'' \emph{IEEE Transactions on Robotics}, 2024.

\bibitem{cai2024pietra}
X.~Cai, J.~Queeney, T.~Xu, A.~Datar, C.~Pan, M.~Miller, A.~Flather, P.~R. Osteen, N.~Roy, X.~Xiao \emph{et~al.}, ``Pietra: Physics-informed evidential learning for traversing out-of-distribution terrain,'' \emph{arXiv preprint arXiv:2409.03005}, 2024.

\bibitem{seo2023learning}
J.~Seo, S.~Sim, and I.~Shim, ``Learning off-road terrain traversability with self-supervisions only,'' \emph{IEEE Robotics and Automation Letters}, vol.~8, no.~8, pp. 4617--4624, 2023.

\bibitem{jung2024v}
S.~Jung, J.~Lee, X.~Meng, B.~Boots, and A.~Lambert, ``{V-STRONG}: Visual self-supervised traversability learning for off-road navigation,'' in \emph{2024 IEEE International Conference on Robotics and Automation (ICRA)}.\hskip 1em plus 0.5em minus 0.4em\relax IEEE, 2024, pp. 1766--1773.

\bibitem{xiao2022learning}
X.~Xiao, T.~Zhang, K.~M. Choromanski, T.-W.~E. Lee, A.~Francis, J.~Varley, S.~Tu, S.~Singh, P.~Xu, F.~Xia, S.~M. Persson, L.~Takayama, R.~Frostig, J.~Tan, C.~Parada, and V.~Sindhwani, ``Learning model predictive controllers with real-time attention for real-world navigation,'' in \emph{Conference on robot learning}.\hskip 1em plus 0.5em minus 0.4em\relax PMLR, 2022.

\bibitem{pan2024traverse}
C.~Pan, A.~Datar, A.~Pokhrel, M.~Choulas, M.~Nazeri, and X.~Xiao, ``Traverse the non-traversable: Estimating traversability for wheeled mobility on vertically challenging terrain,'' \emph{arXiv preprint arXiv:2409.17479}, 2024.

\bibitem{xu2023benchmarking}
Z.~Xu, B.~Liu, X.~Xiao, A.~Nair, and P.~Stone, ``Benchmarking reinforcement learning techniques for autonomous navigation,'' in \emph{2023 IEEE International Conference on Robotics and Automation (ICRA)}.\hskip 1em plus 0.5em minus 0.4em\relax IEEE, 2023, pp. 9224--9230.

\bibitem{xu2021machine}
Z.~Xu, X.~Xiao, G.~Warnell, A.~Nair, and P.~Stone, ``Machine learning methods for local motion planning: A study of end-to-end vs. parameter learning,'' in \emph{2021 IEEE International Symposium on Safety, Security, and Rescue Robotics (SSRR)}.\hskip 1em plus 0.5em minus 0.4em\relax IEEE, 2021, pp. 217--222.

\bibitem{xu2024dexterous}
Z.~Xu, A.~H. Raj, X.~Xiao, and P.~Stone, ``Dexterous legged locomotion in confined 3d spaces with reinforcement learning,'' in \emph{2024 IEEE International Conference on Robotics and Automation (ICRA)}.\hskip 1em plus 0.5em minus 0.4em\relax IEEE, 2024.

\bibitem{karnan2022socially}
H.~Karnan, A.~Nair, X.~Xiao, G.~Warnell, S.~Pirk, A.~Toshev, J.~Hart, J.~Biswas, and P.~Stone, ``Socially compliant navigation dataset (scand): A large-scale dataset of demonstrations for social navigation,'' \emph{IEEE Robotics and Automation Letters}, vol.~7, no.~4, pp. 11\,807--11\,814, 2022.

\bibitem{nguyen2023toward}
D.~M. Nguyen, M.~Nazeri, A.~Payandeh, A.~Datar, and X.~Xiao, ``Toward human-like social robot navigation: A large-scale, multi-modal, social human navigation dataset,'' in \emph{2023 IEEE/RSJ International Conference on Intelligent Robots and Systems (IROS)}.\hskip 1em plus 0.5em minus 0.4em\relax IEEE, 2023, pp. 7442--7447.

\bibitem{karnan2022voila}
H.~Karnan, G.~Warnell, X.~Xiao, and P.~Stone, ``Voila: Visual-observation-only imitation learning for autonomous navigation,'' in \emph{2022 International Conference on Robotics and Automation (ICRA)}.\hskip 1em plus 0.5em minus 0.4em\relax IEEE, 2022, pp. 2497--2503.

\bibitem{xiao2021toward}
X.~Xiao, B.~Liu, G.~Warnell, and P.~Stone, ``Toward agile maneuvers in highly constrained spaces: Learning from hallucination,'' \emph{IEEE Robotics and Automation Letters}, vol.~6, no.~2, pp. 1503--1510, 2021.

\bibitem{xiao2021agile}
X.~Xiao, B.~Liu, and P.~Stone, ``Agile robot navigation through hallucinated learning and sober deployment,'' in \emph{2021 IEEE international conference on robotics and automation (ICRA)}.\hskip 1em plus 0.5em minus 0.4em\relax IEEE, 2021, pp. 7316--7322.

\bibitem{wang2021agile}
Z.~Wang, X.~Xiao, A.~J. Nettekoven, K.~Umasankar, A.~Singh, S.~Bommakanti, U.~Topcu, and P.~Stone, ``From agile ground to aerial navigation: Learning from learned hallucination,'' in \emph{2021 IEEE/RSJ International Conference on Intelligent Robots and Systems (IROS)}.\hskip 1em plus 0.5em minus 0.4em\relax IEEE, 2021, pp. 148--153.

\bibitem{ghani2024dyna}
S.~A. Ghani, Z.~Wang, P.~Stone, and X.~Xiao, ``Dyna-lflh: Learning agile navigation in dynamic environments from learned hallucination,'' \emph{arXiv preprint arXiv:2403.17231}, 2024.

\bibitem{wang2024grounded}
L.~Wang, Z.~Xu, P.~Stone, and X.~Xiao, ``Grounded curriculum learning,'' \emph{arXiv preprint arXiv:2409.19816}, 2024.

\bibitem{miki2022elevation}
T.~Miki, L.~Wellhausen, R.~Grandia, F.~Jenelten, T.~Homberger, and M.~Hutter, ``Elevation mapping for locomotion and navigation using gpu,'' in \emph{2022 IEEE/RSJ International Conference on Intelligent Robots and Systems (IROS)}.\hskip 1em plus 0.5em minus 0.4em\relax IEEE, 2022, pp. 2273--2280.

\end{thebibliography}

\end{document}